# Electronic Health Records-Based Data-Driven Diabetes Knowledge Unveiling and Risk Prognosis

Huadong Pang[1,*,†], Li Zhou[2,†], Yiping Dong[3], Peiyuan Chen[4], Dian Gu[5], Tianyi Lyu[6], Hansong Zhang[7]

[1] Georgia Institute of Technology, USA
[2] Faculty of Management, McGill University, Montreal, QC, Canada, H3B0C7
[3] Department of Mechanical Engineering, Carnegie Mellon University, Pittsburgh, PA, 15213, United States
[4] School of Electrical Engineering and Computer Science, Oregon State University, Corvallis, Oregon, 97333, US
[5] University of Pennsylvania, USA
[6] College of Engineering, Northeastern University, Boston, MA 02115, USA
[7] Department of Electrical and Computer Engineering, University of California, San Diego, La Jolla, 92037, US
* Corresponding author: Huadong Pang (pang_huadong0811@outlook.com )
†These authors contributed equally to this work.

*Abstract*: In the healthcare sector, the application of deep learning technologies has revolutionized data analysis and disease forecasting. This is particularly evident in the field of diabetes, where the deep analysis of Electronic Health Records (EHR) has unlocked new opportunities for early detection and effective intervention strategies. Our research presents an innovative model that synergizes the capabilities of Bidirectional Long Short-Term Memory Networks-Conditional Random Field (BiLSTM-CRF) with a fusion of XGBoost and Logistic Regression. This model is designed to enhance the accuracy of diabetes risk prediction by conducting an in-depth analysis of electronic medical records data. The first phase of our approach involves employing BiLSTM-CRF to delve into the temporal characteristics and latent patterns present in EHR data. This method effectively uncovers the progression trends of diabetes, which are often hidden in the complex data structures of medical records. The second phase leverages the combined strength of XGBoost and Logistic Regression to classify these extracted features and evaluate associated risks. This dual approach facilitates a more nuanced and precise prediction of diabetes, outperforming traditional models, particularly in handling multifaceted and nonlinear medical datasets. Our research demonstrates a notable advancement in diabetes prediction over traditional methods, showcasing the effectiveness of our combined BiLSTM-CRF, XGBoost, and Logistic Regression model. This study highlights the value of data-driven strategies in clinical decision-making, equipping healthcare professionals with precise tools for early detection and intervention. By enabling personalized treatment and timely care, our approach signifies progress in incorporating advanced analytics in healthcare, potentially improving outcomes for diabetes and other chronic conditions.

---

## 1. Introduction

The rapid advancement of medical information technology, especially the widespread use of Electronic Health Records (EHR), has established data-driven approaches as a cornerstone in the field of healthcare big data[1]. Since 2013, countries like the United States and the United Kingdom have significantly increased their investment in medical big data applications, indicating a strong commitment to this field [2]. China has also introduced policies in 2015 and 2016 to foster the development of healthcare big data [3, 4]. This data encompasses a wide range, including personal health information, hospital clinical data, genetic data, and epidemiological data related to disease prevention and control [5, 6].

Diabetes, a major global public health issue, necessitates early diagnosis and effective management to reduce complications and improve patient quality of life[7]. Utilizing the rich data in EHRs, data-driven methods can more accurately predict diabetes risk and offer personalized treatment plans [8, 9]. Moreover, the analysis of healthcare big data helps reduce medical costs, enhance service quality, and supports the formulation of public health policies [10]. The analysis of EHR, which includes patient history, treatment records, and clinical texts, faces many challenges. These include handling unstructured data, assessing data quality, and choosing appropriate analysis methods [11-13]. Addressing these challenges requires more advanced and integrated approaches, particularly those leveraging the synergy of Natural Language Processing (NLP) and machine learning techniques, which can provide more precise insights into patient health and disease risks[14].

In the realm of healthcare, the increasing prevalence of Electronic Health Records (EHR) has turned the deep mining of these data into a key aspect of enhancing the quality of medical services. Particularly, the integration of Natural Language Processing (NLP) and deep learning technologies has opened up new possibilities for extracting essential information from vast amounts of unstructured electronic medical records[15-18]. Against this backdrop, models such as Logistic Regression, Decision Trees, Random Forest, Convolutional Neural Networks (CNN), and Long Short-Term Memory Networks (LSTM) play a significant role in the analysis of EHR data[19]. While traditional models like Logistic Regression and Decision Trees are useful for simpler tasks, deep learning models such as LSTM are more suitable for analyzing unstructured, sequential, or time-series data found in EHRs. Logistic Regression is popular for its simplicity and ease of implementation, is widely used in basic classification tasks, yet shows limitations in handling complex or non-linear relationships [20]. Decision Trees are favored for their interpretability but are prone to overfitting[21]. Random Forest, an ensemble of Decision Trees, enhances model robustness but at the cost of increased model complexity. CNN excels in text classification, especially with spatially structured data, but demands high computational resources [22]. LSTM stands out in processing time-series data due to its ability to handle long-term dependencies, albeit at the cost of model complexity and high training overhead[23, 24].

Given the complexity and unstructured nature of electronic medical records, this article proposes a hybrid method that integrates Natural Language Processing (NLP) for text preprocessing with deep learning models like BiLSTM-CRF for enhanced entity recognition and feature extraction. This approach combines the strengths of deep learning for analyzing complex clinical texts and machine learning models such as XGBoost for accurate diabetes risk prediction, thus improving predictive performance. The integration of these techniques aims to enhance the accuracy of medical information extraction from EHRs, support clinical decision-making, and advance personalized medicine.

This study has made contributions and holds significant importance in the following aspects: We developed a system that leverages natural language processing and machine learning techniques to predict diabetes risk accurately. This is crucial for early diagnosis and intervention in diabetes, ultimately contributing to improved patient health. In addition, our approach can extract valuable medical information and knowledge from electronic health records, including disease features and symptom descriptions. This holds potential value for medical research and decision support, fostering advancements in the healthcare domain. The outcomes of this research lay the foundation for future clinical decision support systems that can aid healthcare professionals in better understanding patient risks and taking appropriate measures. Furthermore, it provides a means for healthcare institutions to enhance medical management and treatment outcomes.

## 2. Literature Review

### 2.1 Overview of Electronic Medical Record Research

The field of Electronic Medical Record (EMR) research globally focuses on utilizing advanced Natural Language Processing (NLP) techniques for clinical text analysis and developing clinical decision support systems. Studies typically employ machine learning models such as Support Vector Machine (SVM), Conditional Random Fields (CRF), and Maximum Entropy (ME) models for named entity recognition and relationship extraction [25-28]. These efforts aim to efficiently and accurately extract useful information from complex clinical texts. Notably, neural network technologies, especially in addressing specific clinical issues like twin fetal weight estimation, have started to be applied and have shown exceptional performance. In recent years, deep learning-based methods have gained popularity for clinical text analysis, further advancing the effectiveness of predictive models in healthcare settings. The development of clinical decision support systems aims to enhance diagnostic accuracy through the analysis of EMR data, marking a significant contribution to the future improvement of healthcare service quality.

In China, EMR research, especially in Chinese language, commenced relatively late but has made considerable progress, particularly in the areas of EMR structural processing and the application of NLP technologies[29]. Research includes not only structural processing of EMRs and summarization of discharge notes but also extends to comprehensive analysis using word segmentation and part-of-speech tagging models. The application of Conditional Random Fields and deep learning methods in entity and relation extraction indicates that Chinese EMR research is gradually aligning with international studies. Despite the foundation provided by resources such as the Chinese Unified Medical Language System (CUMLS) and Traditional Chinese Medical Language System (TCMLS), the relative scarcity of publicly available datasets and biomedical language resources in Chinese poses challenges for further research advancement [30].

Additionally, there has been a shift towards applying machine learning models, such as XGBoost, and advanced techniques like deep learning in recent EMR research to address these challenges more effectively. For example, recent research on fault detection using an improved deep forest approach for modular reconfigurable flying arrays has demonstrated significant results in predictive tasks[31]. This highlights the growing application of ensemble learning methods in medical prediction tasks. Similarly, other studies explore the use of variational autoencoders for time series prediction, which could be applicable in predicting patient outcomes from medical records[32]. These recent works show the applicability of advanced machine learning techniques to healthcare-related challenges, emphasizing the relevance of our methodology.

Overall, the field of EMR research is rapidly evolving globally, achieving technological innovations and increasingly playing a vital role in medical practice. Despite the challenges, the field holds a promising future, particularly in enhancing the accuracy of clinical decision-making and realizing personalized medicine.

## 2.2 Theories and Methods of Data Mining

In the realm of data mining, particularly concerning the analysis of electronic medical records, several key algorithms and methods have been widely applied. These include K-means clustering, Random Forest, Bayesian Theorem, Support Vector Machine (SVM), GBDT, Logistic Regression (LR), and Decision Trees. Each of these methods brings its unique strengths and applicability to handling medical data.

LR is a linear classification method based on the assumption that the conditional distribution P(y|x) is a Bernoulli distribution. It is particularly effective for binary classification problems. The predicted values returned by logistic regression are probability values in the range of [0,1][33].Decision Trees are a type of tree-like classifier where an input sample enters the tree at the root node and is categorized into a class based on feature conditions. Decision Trees are intuitive and can handle both continuous and discrete features. However, they might produce a higher error rate for discrete features with many values and can be complex for continuous features[34-36]. K-means is a popular clustering analysis method in machine learning. Its core idea revolves around the assumption that similar data samples exist in clusters. By selecting k initial centers and assigning each data sample to the nearest center, the algorithm iteratively recalculates center points until convergence. K-means has been extensively used in scenarios like species classification and customer segmentation[37].

Random Forest, initially proposed by Leo Breiman and Adele Cutler, combines bagging and random feature selection [37, 38]. It builds multiple decision trees during training and combines their outputs for final predictions. This approach works effectively for both classification and regression tasks, and is also efficient in ranking feature importance. It evaluates the significance of a feature by comparing the increase in the prediction error when the feature values are permuted in the out-of-bag data. Random Forest is widely used in healthcare applications, particularly in disease prediction, risk stratification, and feature selection.

The Bayesian Theorem, despite its simplicity, remains effective even with smaller datasets. It can incrementally build models and handle multi-class problems, making it particularly suitable for datasets where the amount of data is limited[39]. SVM is a popular supervised learning model commonly used for data analysis and pattern recognition tasks. SVM constructs a non-probabilistic linear model to classify data into two categories and can also perform effectively in non-linear classification tasks through kernel tricks[40].SVM has been applied successfully in healthcare, especially for diagnostic classification tasks, such as identifying disease presence and predicting patient outcomes.

Gradient Boosting Decision Tree (GBDT) is a widely-used machine learning algorithm that is effective for both classification and regression tasks. It combines multiple classification or regression trees in an iterative manner, focusing on minimizing the loss function rapidly and efficiently[41]. GBDT is highly effective in medical data analysis, particularly in patient risk prediction and clinical decision support, where it helps predict outcomes like disease progression and treatment effectiveness.

In addition to traditional machine learning algorithms, deep learning methods such as neural networks, Convolutional Neural Networks (CNNs), and Recurrent Neural Networks (RNNs) are increasingly used for analyzing unstructured healthcare data. For example, CNNs have demonstrated outstanding performance in analyzing medical images, while RNNs are particularly effective in analyzing time-series medical data, such as patient monitoring systems and electronic health records[42, 43]. These deep learning methods have shown significant potential in improving diagnostic accuracy and predicting disease outcomes.

Overall, the combination of traditional machine learning algorithms and advanced deep learning models has significantly advanced the field of medical data analysis. These techniques, applied to electronic health records (EHR), have the potential to revolutionize clinical decision-making, providing healthcare professionals with powerful tools for accurate diagnosis, personalized treatment plans, and early disease detection. As computational power and data availability continue to grow, the integration of these methodologies will further enhance the precision and reliability of predictive models in healthcare.

## 3. Methodology

### 3.1 Overview

The experimental design aims to develop a diabetes risk prediction model by integrating BiLSTM-CRF, Extreme Gradient Boosting (XGBoost), and Logistic Regression. The primary goal is to utilize NLP and machine learning technologies to extract knowledge related to diabetes and predict diabetes risk using Electronic Health Records (EHR).The design involves several key components (Fig.1):

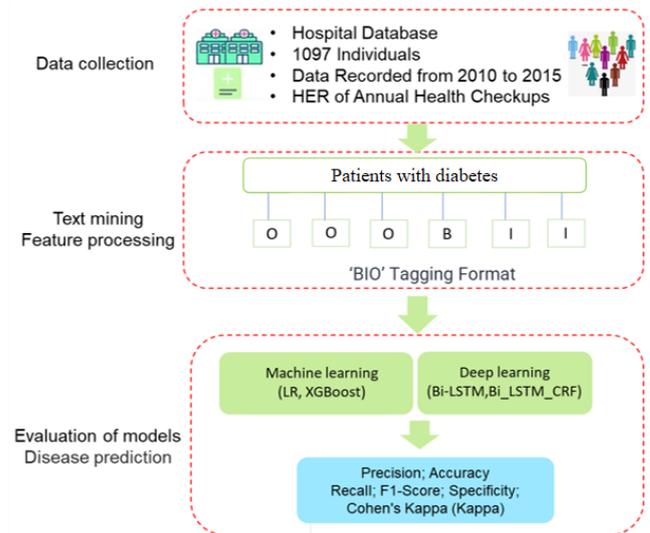

**Fig.1.** Overview of our framework.

(1) Data Preparation: This component focuses on preprocessing text data within EHR. It includes cleaning and standardizing the text, as well as extracting key medical information features such as disease names, symptoms, and treatment methods. The emphasis here is on transforming unstructured text data into a structured format for easier machine processing.

(2) Deep Text Analysis: At this stage, the BiLSTM-CRF model is utilized for in-depth text analysis. The BiLSTM (Bidirectional Long Short-Term Memory) handles text sequences to capture contextual information. The CRF (Conditional Random Field) layer aims to enhance the accuracy of entity recognition. This process effectively extracts key medical entities, including disease features and symptom descriptions, from complex texts.

(3) Feature Integration and Risk Prediction Model Construction: Here, features extracted from texts are merged with other structured data from EHR, such as demographic characteristics of patients and lab results, to form a comprehensive feature set. This set serves as the basis for constructing diabetes risk prediction models using XGBoost and Logistic Regression. XGBoost is chosen for its capability to handle large datasets and capture complex relationships between features, while Logistic Regression is selected for its efficacy in classification and probability estimation.

(4) Model Evaluation and Validation: To thoroughly evaluate the model's performance, we utilize techniques like cross-validation. Key metrics used for this assessment include accuracy, recall, and F1 score. Additionally, ensemble learning methods are considered to combine the predictive results from the XGBoost and Logistic Regression models, enhancing the overall predictive accuracy and robustness.
(5) Final Prediction and Application: Finally, this combined model is used for the ultimate prediction of diabetes risk. This prediction aids not only in early diagnosis of diabetes but also supports clinical decision-making, thereby improving disease management and treatment outcomes.

This experimental design integrates the strengths of NLP and machine learning to create a system capable of accurately predicting diabetes risk and extracting valuable knowledge from EHR. With ongoing optimization and retraining, this system could significantly contribute to clinical decision support and enhance disease management strategies.

## 3.2 BiLSTM

The BiLSTM model is an enhanced Recurrent Neural Network (RNN) that improves sequence learning by analyzing data in both directions, making it particularly effective for time-series tasks like EHR analysis. A standard LSTM model processes data in a forward direction, capturing past information. In contrast, the BiLSTM processes data in both forward and backward directions, capturing both past and future context, which is particularly beneficial for complex sequential data like medical records (Zheng et al., 2021). The core functionality of LSTM lies in its ability to overcome the vanishing gradient problem common in traditional RNNs. This is achieved through its unique structure comprising three types of gates: forget gate; output gate and input gate. The BiLSTM combines two separate LSTMs: one taking the input in a forward direction, and the other in a backward direction. The outputs of these two LSTMs are then combined to form the final output. The typical formulae for an LSTM unit are as follows, and these are applied in both directions in a BiLSTM:

Forget Gate $f_t$: Determines what information to discard from the cell state.

$$f_t = \sigma(W_f \cdot [h_{t-1}, x_t] + b_f) \qquad (1)$$

where $f_t$ controls how much of the previous cell state $C_{t-1}$ should be retained.

Input Gate $i_t$: Updates the cell state with new information.

$$i_t \& = \sigma(W_i \cdot [h_{t-1}, x_t] + b_i) \qquad (2)$$

$$\tilde{C}_t \& = \tanh(W_C \cdot [h_{t-1}, x_t] + b_C) \qquad (3)$$

$$C_t = f_t * C_{t-1} + i_t * \tilde{C}_t \qquad (4)$$

where $i_t$ dictates how much of the new information will be stored in the cell state, and $\tilde{C}_t$ is the new candidate information. The cell state $C_t$ is updated based on the forget gate and input gate.

Output Gate $o_t$ Decides what the next hidden state should be.

$$\begin{aligned} o_t &= \sigma(W_o \cdot [h_{t-1}, x_t] + b_o) \\ h_t &= o_t * \tanh(C_t) \end{aligned} \qquad (5)$$

The output gate determines how much of the cell state $C_t$ information is allowed to pass into the hidden state $h_t$.

σ: Sigmoid activation function, used for gating. tanh: Hyperbolic tangent activation function, used for normalizing data. W and b: Weights and biases, respectively, for the forget gate, input gate, candidate memory cell, and output gate. $[h_{t-1}, x_t]$: Concatenation of the current input $x_t$ and the previous hidden state.

In a BiLSTM(Fig 2.), the outputs from the forward and backward LSTM layers are combined at each time step. The combination method can vary depending on the specific application or architecture, but typically it involves concatenating or summing the hidden states from both directions. The hidden states $h_t^{\text{forward}}$ and $h_t^{\text{backward}}$ from both LSTM layers are concatenated to form a combined hidden state for each time step.

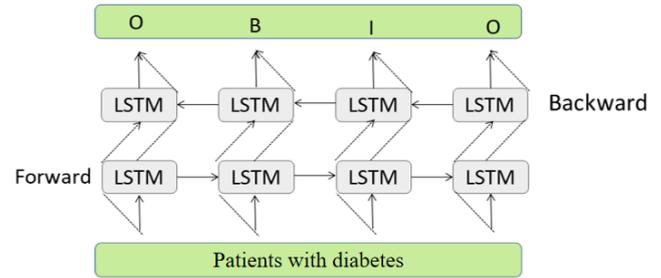

**Fig.2.** BiLSTM model structure.

## 3.3 CRF

The Conditional Random Field (CRF) model is a probabilistic framework used for labeling and segmenting sequential data, particularly useful in natural language processing tasks such as part-of-speech tagging, named entity recognition, and information extraction. Unlike models that make independent assumptions for each data point in a sequence, CRFs consider the context and dependencies between neighboring data points, making them effective for complex sequence modeling tasks.

The basic principle of a CRF is to model the conditional probability of an output sequence (like labels or tags) given an input sequence (such as words in a sentence). Unlike Hidden Markov Models (HMMs), CRFs do not assume independence among the input features, allowing them to capture more complex relationships. The CRF framework is particularly well-suited for tasks where context plays a crucial role in determining the appropriate label for each element in the input sequence.

In mathematical terms, a linear-chain CRF models the conditional probability of a label sequence Y given an input sequence X as follows:

$$P(Y|X) = \frac{1}{Z(X)} \exp\left(\sum_{t=1}^{T} \sum_{k=1}^{K} \lambda_k f_k(y_{t-1}, y_t, X, t)\right) \quad (6)$$

Here, $f_k$ are feature functions that describe the relationship between the labels and the input sequence, $\lambda_k$ are the weights associated with each feature function, T is the length of the sequence, and Z(X) is the normalization factor ensuring that the probabilities sum up to one.

In a CRF model used for diabetes risk prediction, for example, the model could analyze a sequence of medical data points (like lab results over time) and assign a risk level or category to each point based on both the individual data point and its context within the overall sequence. This approach allows for a nuanced understanding of how different medical factors and their progression over time might contribute to the patient's overall risk of developing diabetes.

By leveraging the strengths of CRFs in handling sequential and contextual data, this approach to diabetes risk prediction would offer a sophisticated tool capable of capturing the intricate patterns and relationships inherent in medical data, leading to potentially more accurate and insightful predictions (Fig 3).

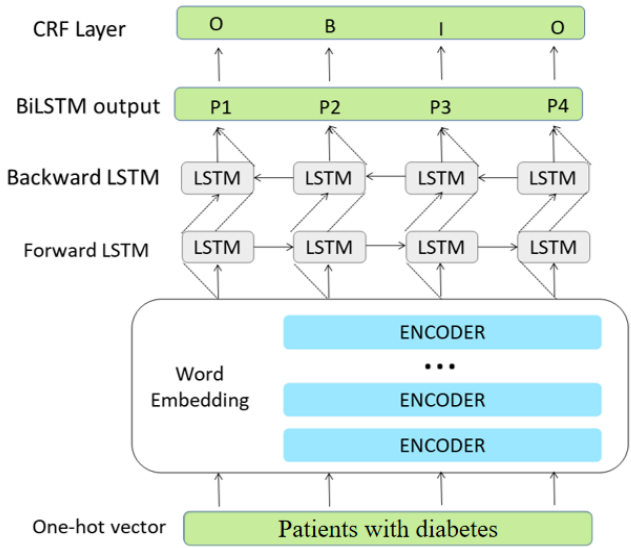

**Fig.3.** BiLSTM-CRF model structure.

## 3.4 GBoost

XGBoost is a highly efficient and widely used machine learning algorithm, particularly effective for structured data [44]. It is an ensemble learning method based on decision trees, primarily used for classification and regression tasks. The core principle of XGBoost involves combining multiple weak learners, typically decision trees, into a strong learner. The fundamental idea behind XGBoost, as a boosting type algorithm, is to sequentially build models, each aiming to reduce the residuals (or errors) of all the previous models. In XGBoost, this is achieved through gradient boosting, where the model is enhanced by minimizing the gradient of the residuals from the previous model. In the context of complex datasets, especially those with nonlinear relationships and intricate interactive effects, XGBoost provides highly accurate solutions. It creates multiple decision trees sequentially and merges their outputs to enhance the overall prediction's accuracy and reliability. In the diabetes risk prediction model you mentioned, XGBoost can be utilized either for feature selection or as an integral part of the predictive model, enhancing the overall performance.

At the heart of XGBoost is the following optimization problem:

$$\text{Obj} = \sum_{i=1}^{n} l(y_i, \hat{y}_i) + \sum_{k=1}^{K} \Omega(f_k) \quad (7)$$

Here, $l(y_i, \hat{y}_i)$ is the loss function, measuring the discrepancy between the predicted values $\hat{y}_i$ and the actual values $y_i$. $\Omega(f_k)$ is a regularization term that penalizes overly complex models to avoid overfitting, typically including the number and depth of the trees. The contribution of each tree is calculated as follows:

$$\hat{y}_i^{(t)} = \hat{y}_i^{(t-1)} + \eta \cdot f_t(x_i) \quad (8)$$

Where $\hat{y}_i^{(t)}$ is the prediction after the t-th round of iteration, η is the learning rate controlling the contribution of each tree to the final prediction, and $f_t(x_i)$ is the prediction of the t-th tree for sample $x_i$. XGBoost iteratively adds trees, each time attempting to reduce the residuals from the previous round, thereby gradually enhancing the model's predictive capability. By combining regularization with multiple iterations, XGBoost effectively prevents overfitting while providing accurate predictive results.

## 4. Experiments
### 4.1 Experimental design

The goal of this experiment is to apply natural language processing and machine learning techniques to extract diabetes-related insights from EHR data and predict diabetes risk. The main goal is to develop a system capable of accurately predicting diabetes risk while extracting valuable insights from EHR data. This system has the potential to greatly enhance clinical decision support and improve disease management practices.

4.1.1 Data Preprocessing
The initial step of our experiment involves data input and preprocessing. We utilize raw textual data from Electronic Health Records (EHR) as our input. Prior to processing, we perform data cleaning to remove irrelevant characters and standardize medical terminology, ensuring data consistency. Additionally, we employ entity recognition using the 'BIO' (Begin, Inside, Outside) tagging format for preprocessing. We identify and label medical entities in the text, such as diseases, symptoms, and treatment methods. We use Natural Language Processing (NLP) techniques to perform this task, marking the beginning of each entity as 'B' (Begin), the middle part as 'I' (Inside), and non-entity parts as 'O' (Outside). Subsequently, we use the preprocessed text data for feature extraction. In the feature extraction phase, we use the Bag of Words model with a vocabulary size of 10,000 and Word Embeddings with a vector dimensionality of 300. This process results in a structured feature dataset where each sample contains text information represented as numerical vectors along with 'BIO' tags related to entity recognition. To address the issue of limited data in our dataset and improve model generalization, we applied data augmentation techniques such as SMOTE (Synthetic Minority Over-sampling Technique) to balance class distribution and prevent overfitting during training.

4.1.2 Deep Text Analysis with BiLSTM-CRF
Moving on to the deep text analysis phase, we employ the BiLSTM-CRF model. This model takes as input the structured feature dataset obtained earlier. Internally, the model utilizes bidirectional Long Short-Term Memory networks (BiLSTM) to process text sequences, capturing contextual information within the text. Furthermore, a Conditional Random Field (CRF) layer is incorporated to enhance entity recognition accuracy, converting tag sequences into optimal entity label sequences. During training, we set the following parameters: a learning rate of 0.001, a batch size of 32, LSTM layer units of 128, and a BiLSTM layer with 2 layers. We also employ the Adam optimization algorithm for model training.

4.1.3 Risk Prediction Model Construction
Subsequently, we proceed to feature fusion and the construction of risk prediction models. In this step, we combine the output tag sequences from the BiLSTM-CRF model with other structured data from the EHR, forming a comprehensive feature set. We then build two independent risk prediction models: the XGBoost model and the Logistic Regression model. For the XGBoost model, we set specific parameters, including a learning rate of 0.1, a maximum tree depth of 5, a minimum child weight of 1, subsampling with a ratio of 0.8, and column subsampling with a ratio of 0.8. For the Logistic Regression model, we set the regularization parameter (C) to 0.1 and utilize the L-BFGS optimization algorithm.

4.1.4 Model Evaluation and Metrics
In our study, we used BiLSTM-CRF, XGBoost, and Logistic Regression models for training and evaluating data. We carefully set the training data ratio, input data, and evaluation methods for each model.
For the BiLSTM-CRF model, we split the dataset into an 80% training set and a 20% test set. This means that 80% of the data is used to train the model, while the remaining 20% is reserved for testing and evaluating its performance. To address potential overfitting in the BiLSTM-CRF model, we implemented early stopping based on validation loss and applied dropout regularization with a rate of 0.5 during training. With the XGBoost and Logistic Regression models, the input data usually comes in the form of structured features from electronic health records, including patient physiological parameters and medical history. These models might not directly process raw textual data but use features that have been converted into numerical form. To ensure that our models generalize well across different populations and datasets, we adopt several strategies to improve model robustness. First, we employ transfer learning by fine-tuning pre-trained models like Med-BERT, which has been trained on a large, diverse corpus of medical data. Fine-tuning this model on our specific dataset allows the model to leverage broader knowledge, reducing the risk of overfitting to the specific characteristics of our training data. Additionally, to enhance the model's ability to generalize to diverse populations, we ensure that our training set includes a variety of demographic groups, healthcare conditions, and medical practices. By doing so, the model learns more generalized features that are applicable across different sub-populations. To further mitigate overfitting and ensure robust model performance, we applied 5-fold cross-validation for both XGBoost and Logistic Regression models. This helps to assess how well the models generalize by evaluating them on multiple different data subsets. Additionally, Logistic Regression was regularized using an L2 penalty to prevent overfitting.
For evaluation, in addition to Accuracy, Recall, and F1-Score, we also use other metrics. In classification tasks, each sample in the data has a true class, and the model predicts a class for each sample. Classification evaluation metrics are divided into those for binary classification problems and those for multiclass classification problems. In binary classification, we can categorize samples into four classes based on the Confusion Matrix:
True Positive (TP): Correctly identified positive instances.
True Negative (TN): Correctly identified negative instances.
False Positive (FP): Incorrectly classified as positive, but truly negative.
False Negative (FN): Incorrectly classified as negative, but truly positive.
Accuracy:
$$\text{Accuracy} = \frac{TP+TN}{TP+TN+FP+FN} \quad (9)$$
Precision:
$$\text{Precision} = \frac{TP}{TP+FP} \quad (10)$$
Recall:
$$\text{Recall} = \frac{TP}{TP+FN} \quad (11)$$
F1-Score:
$$F1 - \text{Score} = \frac{2 \cdot \text{Precision} \cdot \text{Recall}}{\text{Precision} + \text{Recall}} \quad (12)$$
Specificity: Specificity is the proportion of True Negatives out of all actual negative instances. It measures the model's ability to correctly identify negative instances.
$$\text{Specificity} = \frac{TN}{TN+FP} \quad (13)$$
Cohen's Kappa (Kappa):
$$\text{Kappa} = \frac{P_o - P_e}{1 - P_e} \quad (14)$$
$P_o$ represents the observed agreement, and $P_e$ represents the expected agreement under random chance.

### 4.2 Dataset

Font: Open Sans; Font size; 10. Paragraph comes content here. Paragraph comes content here.

Our experiment utilized a dataset from a renowned health check center in Beijing, consisting of health records of 5,046 individuals, focusing on data from January 2010 to December 2015. To ensure a comprehensive and standardized dataset, we selectively included individuals with annual health checks and excluded those with either fewer than five years of records or multiple checks per year, narrowing down to 1,097 individuals with 48 features, including gender, age, AST/ALT ratio, proportion, hematocrit, red cell distribution width, mean corpuscular volume, creatine kinase (CK), fasting blood glucose, pupil dilation, lymphocyte percentage, total lymphocyte count, urine bilirubin, urine urobilinogen, urine protein, urinalysis - white blood cells, urinalysis - red blood cells, urea, urine pH, urine glucose, urine ketones, mean corpuscular hemoglobin concentration, mean platelet volume, height, lower limb edema, weight, body mass index (BMI), electrocardiogram (ECG), heart rate, complete blood count - white blood cells, complete blood count - red blood cells, serum aspartate aminotransferase (AST), serum low-density lipoprotein cholesterol (LDL-C), serum triglycerides, serum high-density lipoprotein cholesterol (HDL-C), serum alanine aminotransferase (ALT), serum total cholesterol, platelet distribution width, platelet count, nitrites, waist circumference, fatty liver, neutrophil percentage, total neutrophil count, systolic blood pressure, diastolic blood pressure, and diabetes status.

The preprocessing of this data involved a meticulous process of cleaning and standardization, where irrelevant characters were removed, and medical terminologies were standardized to maintain uniformity. In extracting meaningful features from the rich textual descriptions in the health reports, such as doctor's annotations, diagnostic results, and treatment recommendations, we deployed Natural Language Processing (NLP) techniques. Two primary methods were utilized: the BIO tagging format and Word Embeddings. This intricate process of feature extraction and fusion was pivotal in discerning the complex layers of information embedded in the health reports, playing a critical role in assessing the risk of diabetes, monitoring its progression, and aiding in the development of tailored health recommendations. Our integrated approach significantly bolstered the depth and effectiveness of our analysis, proving crucial in the predictive modeling of diabetes-related outcomes.

### 4.3 Comparison study results and analysis

Table 1. Comparison between different models.

| Model | Accuracy | Precision | Recall | F1 Score | Specificity | Kappa Coefficient |
|---|---|---|---|---|---|---|
| LSTM | 0.70 | 0.6240 | 0.6942 | 0.7005 | 0.6875 | 0.4003 |
| BiLSTM | 0.76 | 0.7828 | 0.6715 | 0.7229 | 0.6962 | 0.5560 |
| CNN-Bi-LSTM[45] | 0.77 | 0.7812 | 0.6815 | 0.7284 | 0.6981 | 0.5624 |
| 3D-CNN-SPP[46] | 0.80 | 0.7981 | 0.6879 | 0.7341 | 0.6997 | 0.5681 |
| Med-BERT[47] | 0.81 | 0.8073 | 0.7982 | 0.7399 | 0.7055 | 0.6182 |
| BiLSTM-CRF | 0.82 | 0.8090 | 0.6968 | 0.7487 | 0.7097 | 0.6219 |

The table 1 displays the performance metrics of various models. The BiLSTM-CRF model achieves the highest accuracy of 82%, followed by Med-BERT at 81%, 3D-CNN-SPP at 80%, and CNN-Bi-LSTM at 77%.

In terms of precision, BiLSTM-CRF leads with 80.90%, followed closely by Med-BERT at 80.73% and BiLSTM at 78.28%. The BiLSTM-CRF model also shows the highest F1 score of 74.87%, outperforming BiLSTM at 72.29% and CNN-Bi-LSTM at 73.41%. In terms of recall, LSTM achieves the highest recall at 79.82%, but sacrifices precision and specificity. BiLSTM-CRF, however, achieves a more balanced performance across precision, recall, and F1 score, demonstrating its superiority in predicting diabetes risk. The Kappa coefficient is highest for BiLSTM-CRF at 0.6219, indicating strong agreement between the model's predictions and the actual outcomes. This suggests that BiLSTM-CRF has minimal overfitting and is highly generalizable. BiLSTM-CRF outperforms all other models across key metrics, demonstrating the effectiveness of combining Bidirectional LSTM for enhanced contextual understanding of medical texts and Conditional Random Fields for optimizing entity recognition. This hybrid approach strengthens the model's ability to handle both structured and unstructured data, which is crucial for accurate diabetes risk prediction. Figure 4 shows the graphical representation of the model comparison.

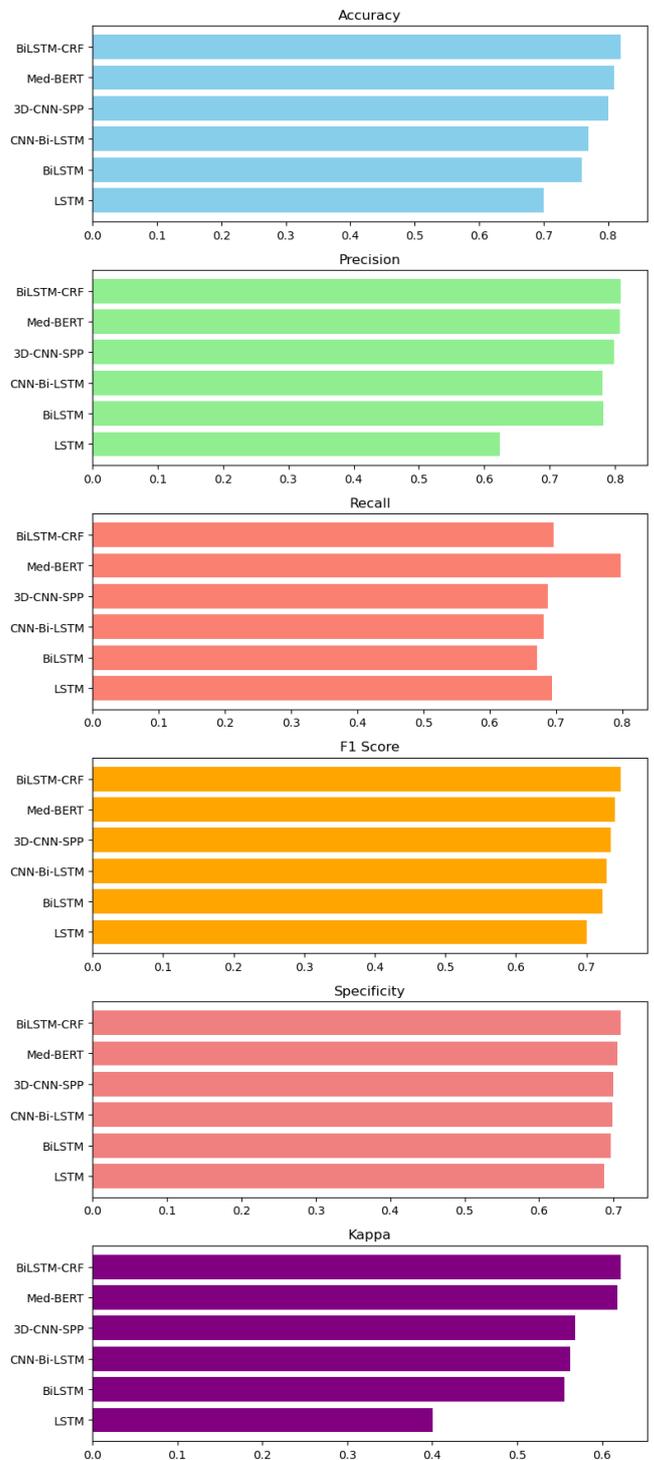

**Fig.4.** Comparison between different models.

Figure 5 depicts the F1 score trajectories of three deep learning models over 100 training epochs. The F1 score, as a harmonic mean of precision and recall, serves as a robust metric for model accuracy, particularly in imbalanced dataset contexts. The LSTM model, represented by the blue line, exhibits a steep initial learning curve that plateaus just above the 0.70. The BiLSTM model, indicated by the orange line, follows a similar trajectory with a marginally higher convergence value, suggesting a slightly better balance between precision and recall. The BiLSTM-CRF model, denoted by the green line, shows a comparable initial rise but fluctuates around the 0.75 level after the 20-epoch mark, indicating the CRF layer's nuanced capability to capture dependencies, resulting in a higher F1 score. All models demonstrate rapid performance improvements within the initial 20 epochs, followed by a deceleration in learning rate and subsequent stabilization around their respective score levels. This pattern of quick initial improvement followed by a plateau is indicative of models reaching their learning capacity with the given training data. The closeness of the curves post the 20-epoch threshold implies similar predictive capabilities across the models, with the CRF-enhanced model potentially offering slight enhancements in performance.

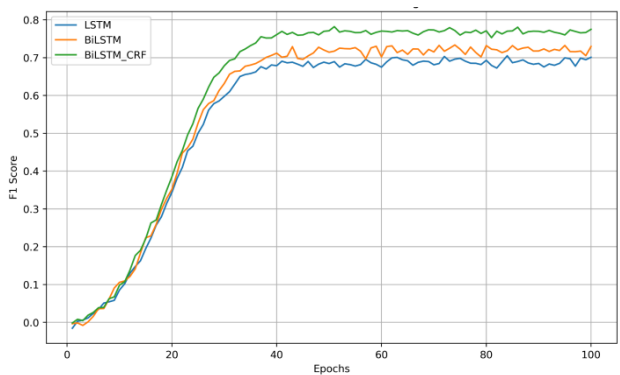

**Fig.5.** Comparison F1 score between different models.

**Table 2.** Comparative Performance Metrics between different models.

| Metric | XGBoost | Logistic Regression | Ensemble |
| --- | --- | --- | --- |
| Accuracy | 0.86 | 0.81 | 0.89 |
| Precision | 0.88 | 0.83 | 0.90 |
| Recall | 0.85 | 0.79 | 0.87 |
| F1 Score | 0.86 | 0.81 | 0.88 |
| AUC | 0.91 | 0.87 | 0.93 |

The table 2 presents the performance metrics of three models – XGBoost, Logistic Regression, and an Ensemble model. The Ensemble model outperforms both XGBoost and Logistic Regression across all key metrics, demonstrating the benefits of combining multiple models. In terms of accuracy, the Ensemble model achieved 89%, surpassing XGBoost at 86% and Logistic Regression at 81%. This highlights the Ensemble model's ability to improve classification accuracy. For precision, the Ensemble model leads with 90%, followed by XGBoost at 88% and Logistic Regression at 83%. This suggests the Ensemble model excels at minimizing false positives. In recall, the Ensemble model is again the best with 87%, compared to XGBoost at 85% and Logistic Regression at 79%. The Ensemble model achieves a better balance between precision and recall.

The F1 score, combining precision and recall, also favors the Ensemble model at 88%, with XGBoost at 86% and Logistic Regression at 81%. This shows the Ensemble model's balanced performance. The AUC, indicating discriminatory power, is highest for the Ensemble model at 93%, followed by XGBoost at 91% and Logistic Regression at 87%. This confirms the Ensemble model's superior ability to distinguish between classes. In conclusion, the Ensemble model consistently outperforms the other models, making it the most effective choice for this predictive task, with superior accuracy, precision, recall, F1 score, and AUC.

## 5. Conclusion

In this study, we aimed to address the challenge of diabetes risk prediction and knowledge discovery from EHR in a data-driven manner. Our primary goal was to accurately predict diabetes risk and extract valuable insights from extensive medical text and structured EHR data. To achieve this, we employed several key methods: first, preprocessing the textual data through text cleaning, standardization, and extraction of medical features. The BiLSTM-CRF model was then used for in-depth text analysis, with BiLSTM capturing contextual information and the CRF layer improving entity recognition accuracy. The extracted text features were integrated with structured data, such as patient demographics and lab results, to create a comprehensive feature set. This formed the basis for diabetes risk prediction models built using XGBoost, which excels at handling large datasets, and Logistic Regression, known for its effective classification and probability estimation.

Despite achieving certain milestones in our research, two primary deficiencies and future prospects are evident: Our experimental dataset is relatively small, consisting of only 1000 clinical electronic health records. To comprehensively train and validate our models, it is essential to expand the dataset to capture a more extensive range of diseases and patient characteristics. Future work should concentrate on data collection and augmentation, possibly through collaborations with multiple healthcare institutions to ensure diversity and representativeness of the data. While we employed deep learning and machine learning models for prediction and knowledge discovery, these models are often considered black-box models, making it challenging to interpret their prediction results. Future research should focus on enhancing the interpretability of the models, enabling healthcare professionals to understand the decision-making process. Additionally, further model optimization is a future direction to improve prediction accuracy and stability. While we focused on structured and textual data, incorporating additional factors such as genetic, environmental, and behavioral data could further enhance model performance and better capture the multifaceted nature of diabetes. In addition, although this study used a public dataset, we acknowledge the ethical concerns regarding privacy, informed consent, and data security with personal health data. Future studies involving private health data should ensure compliance with ethical standards and institutional review board (IRB) approvals. We also suggest exploring privacy-preserving techniques, such as federated learning, to protect sensitive data while enabling meaningful analysis.

In summary, this study's approach harnessed the strengths of Natural Language Processing and machine learning to create a system capable of accurately predicting diabetes risk and extracting valuable insights from EHR. Through ongoing model optimization and retraining, this system could make significant contributions to clinical decision support and disease management, ultimately improving patient outcomes and healthcare efficiency. This research presents an innovative approach to combining data-driven diabetes risk prediction and medical knowledge discovery, holding significant clinical and research significance. Future work should focus on overcoming deficiencies related to data limitations and model interpretability while further advancing this field to benefit both healthcare professionals and patients.

**Conflicts of Interest**

The authors declare that they have no conflicts of interest.

**Author contributions**

Huadong Pang designed the research and supervised the entire project, while Li Zhou handled data collection and preprocessing. Yiping Dong contributed to the development of methods and algorithm improvements. Peiyuan Chen provided support during experimental implementation and validation. Dian Gu assisted in the literature review and contributed to drafting parts of the manuscript. Tianyi Lyu was responsible for statistical analysis of the data and the creation of visual representations of results. Hansong Zhang provided technical support, helped build experimental tools, and participated in refining and revising the manuscript.